\documentclass[runningheads]{llncs}
\usepackage[T1]{fontenc}
\usepackage{graphicx}
\usepackage{subcaption}
\usepackage{amsmath,amssymb,amsfonts}
\usepackage{color}

\usepackage{algorithm}
\usepackage{algpseudocode}
\usepackage{amsmath}
\usepackage{url}
\usepackage[table]{xcolor}

\begin{document}
\title{Sensory Precision Inference for Multimodal Arbitration under Uncertainty}
%
%
\author{Tin Mi\v si\'c\inst{1} \and 
Takato Horii\inst{1,2}}

\authorrunning{T. Mi\v si\'c and T. Horii}

\institute{The University of Osaka, Japan \and
IRCN, The University of Tokyo, Japan \\Correspondence: \email{misic.tin.rtf@ecs.osaka-u.ac.jp}}
\maketitle              
\begin{abstract}
Autonomous agents operating on multisensory data cannot assume that all sensory modalities remain consistently informative. In real environments, sensory streams are frequently corrupted by noise, missing data, or inter-modal incongruence, requiring adaptive arbitration between competing sensory hypotheses. While active inference provides a principled framework for uncertainty-guided inference, the role of dynamically inferred sensory precision in generative multimodal arbitration under sensory conflict remains comparatively underexplored.
We propose a multimodal active inference model in which latent beliefs and modality-specific sensory precisions are jointly updated through iterative free-energy minimization. 
In our proposed model, sensory precision dynamics not only reflect sensory uncertainty but actively shape the evolution of latent beliefs during multimodal conflict. In addition, we introduce a learned prior over sensory precisions that induces structured, class-dependent precision patterns and influences cross-modal inference dynamics.
We evaluate the model using a synthetic multimodal MNIST dataset combining visual, auditory, and tactile representations of digit classes under controlled sensory noise, modality dropout, and inter-modal incongruence. Results show that dynamic precision inference improves reconstruction robustness under corrupted sensory evidence, enables coherent multisensory belief formation from partial observations, and produces stable arbitration between conflicting modalities. Furthermore, learned precision priors generate interpretable precision structures that shape inference dynamics and cross-modal latent structure.
These findings support sensory precision inference as a mechanistic control process for adaptive multimodal belief formation under uncertainty, highlighting precision dynamics as a computational mechanism for robust and interpretable multisensory integration.

\keywords{Cross-modal arbitration \and Sensory precision \and Active inference}
\end{abstract}
\section{Introduction}


Autonomous agents operating in multisensory environments cannot assume that all sensory modalities remain consistently reliable. Real-world sensory streams are frequently degraded by noise, partial sensory loss, or inter-modal incongruence, requiring adaptive arbitration between competing sources of sensory evidence. Biological systems dynamically redistribute inferential reliance across modalities, selectively prioritizing sensory signals that are expected to provide reliable information. Phenomena such as the McGurk effect \cite{mcgurk76} and the Stroop effect \cite{stroop35} further demonstrate the ability of the brain to resolve conflicting sensory evidence through selective modulation of perceptual influence.

\begin{figure}[t]
    \centering
    \includegraphics[width=\linewidth]{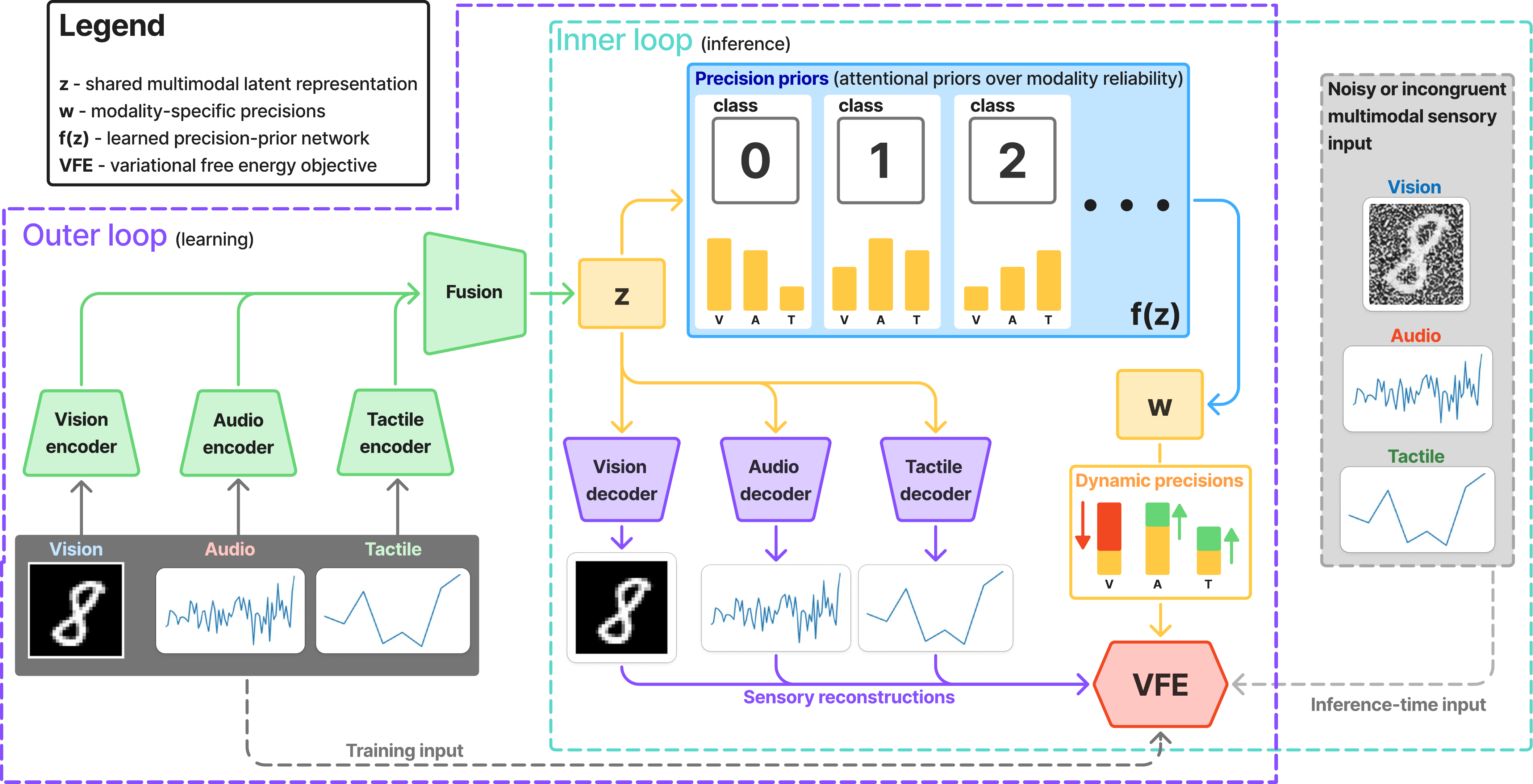}
    \caption{Proposed multimodal precision inference framework. The outer loop performs parameter learning for the encoder, decoder, fusion, and precision-prior networks, while the inner loop performs iterative free-energy minimization over latent beliefs and modality precisions.}
    \label{fig:model}
\end{figure}

Within active inference, precision corresponds to the expected reliability of sensory observations when inferring latent causes \cite{Parr18,Friston09}. Computationally, precision modulates the gain of prediction errors during inference \cite{Friston10}: highly precise sensory signals exert stronger influence on belief updates, whereas low-precision signals are attenuated. In predictive processing accounts, such precision modulation has frequently been associated with attentional allocation \cite{Feldman10}, allowing inferential influence to adapt dynamically according to contextual uncertainty.

Previous work has investigated precision dynamics in spatial attention, active sensing, and policy selection \cite{Feldman10,Misic26,Mirza2019-iq,Parr18,Parr17}. These approaches demonstrate how precision can guide selective sampling and attentional shifts according to the agent's beliefs and environmental context. Priors over sensory precision further shape how inferential influence is distributed before sensory observations are fully resolved, arising from prior preferences or learned expectations regarding environmental structure \cite{Shomstein23,Parvizi-Wayne24,Feldman10,Misic26}. In multimodal settings, latent causes are often associated with distinct modality-specific reliability structures, suggesting that object representations may encode structured expectations regarding the relative informativeness of different sensory modalities \cite{Lynott13,Van_de_Weijer24,Molholm07}.
Under uncertain conditions, such learned precision priors may guide the redistribution of inferential influence across modalities. 

Related challenges have also been explored in robotics and multimodal machine learning. Variational Bayesian approaches to human-robot interaction and multimodal concept learning have incorporated modality-specific weighting and interdependencies between sensory streams \cite{Horii18,Horii21,r25,r26,r29}. 
More broadly, multimodal neural architectures have investigated robustness to noisy or degraded sensory inputs through weighted attention mechanisms and uncertainty-aware encodings \cite{Fronzaglia25,Liu22,Mai24,Gao24,Zhang25,Nakamura23}. However, in many existing approaches modality weighting remains fixed, learned statically during training, or optimized primarily for discriminative objectives. Comparatively little work has examined dynamically inferred modality precision within generative multimodal inference, particularly under sensory conflict, dropout, or inter-modal incongruence. 

To address these limitations, we propose a multimodal active inference framework focused on precision-based attentional control during perceptual inference\footnote{Code, training and experiment details available at: \url{https://github.com/TinMisic/multimodal_attention}}, with the following contributions: 
\begin{itemize}
\item A multimodal active inference architecture that performs joint inference over latent representations and modality-specific sensory precisions, enabling adaptive arbitration under noise, modality dropout, and sensory conflict.
\item A learned class-dependent precision-prior mechanism that encodes structured expectations regarding modality reliability and provides top-down influences on precision allocation during inference.
\item A computational analysis of precision dynamics demonstrating how precision-weighted prediction errors influence latent belief trajectories, suppress unreliable sensory evidence, and support multimodal conflict resolution.
\end{itemize}
The framework is evaluated using controlled experiments on a synthetic multimodal MNIST dataset comprising visual, auditory, and tactile representations.
A graphical overview of the proposed framework is shown in Fig.~\ref{fig:model}.

\section{Proposed Method}
\label{method}
\subsection{Generative Model}

Consider a set of multimodal observations \(\mathbf{o} = \{o^{(1)}, ..., o^{(M)}\}\), where each modality \(m \in \{1,...,M\}\) corresponds to a sensory observation of dimensionality \(d_m\). The model assumes a shared latent representation \(\mathbf{z} \in \mathbb{R}^{D}\), which captures the underlying latent causes of the sensory observations. In addition, modality-specific precision variables \(\mathbf{w} \in \mathbb{R}^{M}\) parameterize the expected reliability of each sensory modality during inference.
The full generative model factorizes as
\begin{equation}
    p_{\theta}(\mathbf{o}^{(1:M)}, \mathbf{z}, \mathbf{w})
    =
    p_{\theta}(\mathbf{z})
    p_{\theta}(\mathbf{w}|\mathbf{z})
    \prod_{m=1}^{M}
    p_{\theta}(\mathbf{o}^{(m)}|\mathbf{z},\mathbf{w}).
\end{equation}
The latent prior over \(\mathbf{z}\) is defined as a standard Gaussian,
\begin{equation}
    p_{\theta}(\mathbf{z})
    =
    \mathcal{N}(\mathbf{z};\mathbf{0},\mathbf{I}).
\end{equation}

To model structured expectations regarding modality reliability, we introduce a latent-dependent prior over sensory precision variables:
\begin{equation}
    p_{\theta}(\mathbf{w}|\mathbf{z})
    =
    \mathcal{N}(\mathbf{w};f_{\theta}(\mathbf{z}),\mathbf{I}),
\end{equation}
where \(f_{\theta}(\mathbf{z})\) is a learned mapping from latent representations to expected modality precision configurations.
Each modality likelihood is modeled as
\begin{equation}
    p_{\theta}(\mathbf{o}^{(m)}|\mathbf{z},\mathbf{w})
    =
    \mathcal{N}(\mathbf{o}^{(m)};g_m(\mathbf{z}),\Pi_m(\mathbf{w})^{-1}),
\end{equation}
where \(g_m(\mathbf{z})\) denotes the modality-specific decoder network.

The precision variables are transformed into positive precision scalars through the exponential function.
The resulting modality precision matrices for each modality are given by
\begin{equation}
    \Pi_m(\mathbf{w})
    =
    \Sigma_m(\mathbf{w})^{-1}
    =
    \frac{\exp(w_m)}{\sigma_{m,o}^2}\mathbf{I}_{d_m},
\end{equation}
with \(\Sigma_m\) being the modality covariance matrices and  \(\sigma_{m,o}^2\) being the modality-specific base-scale observation covariances, which are estimated from training-set statistics and fixed during inference.
Under this formulation, increasing \(w_m\) increases the precision assigned to modality \(m\), thereby increasing the influence of modality-specific prediction errors.

\subsection{Variational Free Energy}

Inference is performed through minimization of variational free energy under a mean-field variational approximation,
\begin{equation}
    q(\mathbf{z},\mathbf{w})
    =
    q(\mathbf{z})q(\mathbf{w}),
\end{equation}
with Gaussian approximate posteriors
\begin{equation}
\begin{aligned}
    q(\mathbf{z})
    & =
    \mathcal{N}(\mathbf{z};\mu_z,\Sigma_z),\\
    q(\mathbf{w})
    & =
    \mathcal{N}(\mathbf{w};\mu_w,\Sigma_w),
\end{aligned}    
\end{equation}
with covariances $\Sigma_z$ and $\Sigma_w$. The variational free energy objective is defined as
\begin{equation}
    \begin{aligned}
        \mathcal{F}[q] & =\mathbb{E}_{q(z,w)}\left[ log~q(\mathbf{z},\mathbf{w}) - log~p_\theta(\mathbf{o},\mathbf{z},\mathbf{w}) \right]\\
        & = \mathbb{E}_{q(z,w)}\left[- log~p_\theta(\mathbf{o}|\mathbf{z},\mathbf{w}) \right] +D_{KL}(q(\mathbf{z})||p(\mathbf{z})) \\
        & + \mathbb{E}_{q(z)}\left[D_{KL}(q(\mathbf{w})||p(\mathbf{w|z}))\right].
    \end{aligned}
\end{equation}
Inference is implemented through iterative optimization of posterior means, yielding a point-estimate approximation of the variational posterior.
Under a point-estimate approximation, posterior covariance terms are treated as constants and omitted from the optimization objective. The resulting free energy depends only on the posterior means:
\begin{equation}
\begin{aligned}
    \mathcal{F}(\mu_z,\mu_w)
    &=
    \frac{1}{2}
    \sum_m
    \varepsilon_m^T
    \Pi_m(\mu_w)
    \varepsilon_m
    +
    \frac{1}{2}
    \log\det(2\pi\Sigma_m(\mu_w)) \\
    &+
    \frac{1}{2}\mu_z^T\mu_z
    +
    \frac{1}{2}
    \varepsilon_w^T
    \varepsilon_w,
\end{aligned}
\end{equation}
where modality prediction errors and precision prediction errors are defined as
\begin{equation}
    \begin{aligned}
        \varepsilon_m
        &=
        o^{(m)}-g_m(\mu_z),\\
        \varepsilon_w
        &=
        \mu_w-f(\mu_z).
    \end{aligned}
\end{equation}

The free energy objective therefore jointly penalizes sensory prediction errors, latent complexity, and deviations from expected precision configurations. 

\subsection{Dynamic Precision Inference}

Beliefs over latent states and sensory precisions are jointly updated through gradient descent on variational free energy:
\begin{equation}
    \begin{aligned}
        \mu_z
        &\leftarrow
        \mu_z
        -
        \eta_z
        \frac{\partial \mathcal{F}}{\partial \mu_z},\\
        \mu_w
        &\leftarrow
        \mu_w
        -
        \eta_w
        \frac{\partial \mathcal{F}}{\partial \mu_w}.
    \end{aligned}
\end{equation}

The latent-state gradient is given by
\begin{equation}
\frac{\partial \mathcal{F}}{\partial \mu_z}
=
\mu_z
-
\sum_{m=1}^{M}
\left(
\frac{\partial g_m(\mu_z)}
{\partial \mu_z}
\right)^T
\Pi_m(\mu_w)
\varepsilon_m
-
\left(
\frac{\partial f(\mu_z)}
{\partial \mu_z}
\right)^T
\varepsilon_w.
\end{equation}
This update consists of three distinct contributions. The first term corresponds to the latent prior, which regularizes latent beliefs toward the origin. The second term corresponds to precision-weighted sensory prediction errors arising from each modality. 
The third term introduces a top-down contribution arising from the learned precision prior \(f(\mathbf{z})\). Through this term, latent representations encode structured expectations regarding modality reliability, allowing precision prediction errors to influence latent-state dynamics. Consequently, latent beliefs are shaped not only by bottom-up sensory prediction errors, but also by learned expectations regarding modality-specific uncertainty.

The precision-state gradient is given by
\begin{equation}
\frac{\partial \mathcal{F}}{\partial \mu_w}
=
\varepsilon_w
+
\frac{1}{2}
\exp(\mu_w)
\odot
\left(
 \varepsilon_m^T
    \Pi_{m}(\mu_w)
    \varepsilon_m
 -
 \frac{d_m}{\exp(\mu_w)}
\right).
\end{equation}
Under this formulation, precisions dynamically adapt according to both sensory prediction error magnitude and latent-dependent precision expectations. Modalities producing large prediction errors tend to receive reduced inferential influence, while modalities consistent with the inferred latent state are preferentially weighted during multimodal arbitration.

\section{Model Implementation and Experimental Design}
\label{experiments}

\subsection{Dataset and Multimodal Setup}
 
The experiments were conducted using a synthetic multimodal dataset constructed from the MNIST handwritten digit dataset consisting of 10 digit classes. The original \(28 \times 28\) grayscale images were used as the visual modality, while additional auditory and tactile modalities were synthetically generated as 64-dimensional and 16-dimensional feature vectors, respectively.

\begin{figure}[t]
    \centering
    \begin{subfigure}[b]{0.2\linewidth}
        \centering
        \includegraphics[width=0.9\linewidth]{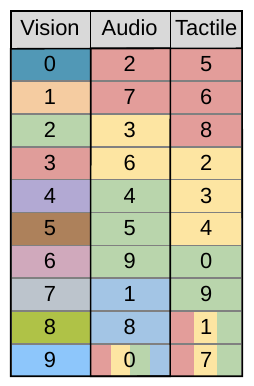}
        \caption{ }
        \label{fig:table}
    \end{subfigure}
    \hfill
    \begin{subfigure}[b]{0.42\linewidth}
        \centering
        \includegraphics[width=\linewidth]{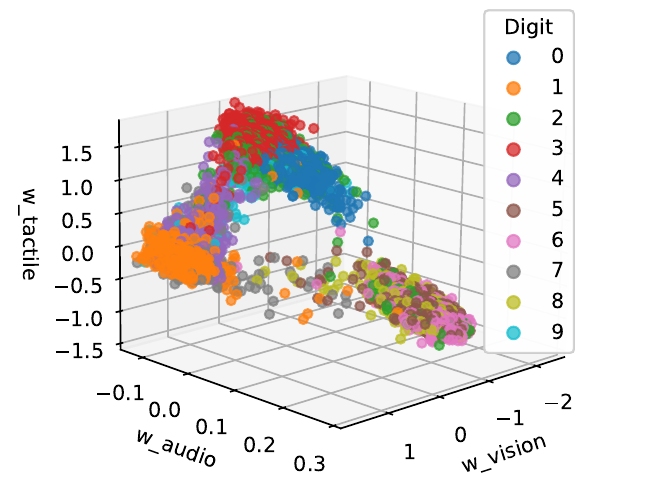}
        \caption{ }
        \label{fig:f(z)}
    \end{subfigure}
    \hfill
    \begin{subfigure}[b]{0.35\linewidth}
        \centering
        \includegraphics[width=\linewidth]{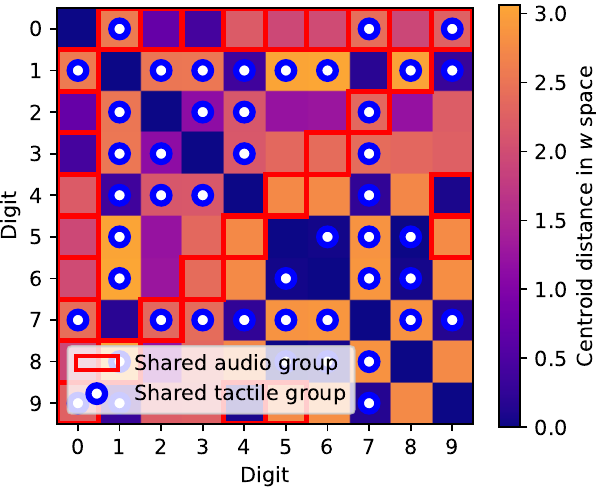}
        \caption{ }
        \label{fig:sep}
    \end{subfigure}
    \caption{
        (a) Audio and tactile class groupings used to introduce modality-specific ambiguity during dataset generation.
        (b) Learned precision-prior mapping \(f(\mathbf{z})\).
        (c) Pairwise distances between class centroids in \(\mathbf{w}\)-space; marked pairs correspond to classes sharing the same audio or tactile group from (a).
        }
    \label{fig:data}
\end{figure}

To induce modality-specific ambiguity and varying reliability structures, the synthetic auditory and tactile modalities were generated from Gaussian distributions centered around randomly initialized representative modality vectors. For the tactile modality, eight digit classes were assigned to three shared modality groups containing three, three, and two digit classes respectively, while the remaining two classes sampled observations stochastically from these existing group distributions. Similarly, the auditory modality contained three shared modality groups of two digit classes each, one group of three classes, and the remaining class sampled randomly from the existing auditory groups. A table with the modality groupings can be seen in Fig.\ref{fig:table}. This construction introduces partial overlap and stochastic ambiguity within individual modalities, preventing perfect unimodal discrimination and encouraging reliance on multimodal integration and adaptive precision inference during belief formation. 

All modality features were normalized to zero mean and unit variance across the dataset. The dataset followed the standard MNIST split consisting of 60,000 training samples and 10,000 test samples, with the training set further divided into 90\% training and 10\% validation subsets.

\subsection{Model Implementation, Training and Inference}

The model consists of modality-specific encoders and decoders together with an iterative free-energy inference process over latent beliefs and modality precisions. 
The visual modality encoder consists of a convolutional neural network operating on \(28 \times 28\) grayscale images, while the auditory and tactile modalities use multilayer perceptrons (MLPs). Individual modality embeddings are concatenated and fused through an MLP fusion network to produce the shared latent representation \(\mathbf{z} \in \mathbb{R}^{32}.\)
The latent representation is then passed to modality-specific decoder networks which reconstruct the sensory observations for each modality. In addition, the latent state \(\mathbf{z}\) is provided as input to the learned precision-prior network \(f(\mathbf{z})\), which predicts the expected modality precision configuration \(\mathbf{w} \in \mathbb{R}^{3}.\)
The latent precision variables \(\mathbf{w}\) determine the weighting of modality-specific prediction errors during iterative inference.
Prediction errors were normalized according to modality dimensionality to prevent high-dimensional modalities from dominating inference dynamics.

Training was performed using a dual-loop optimization procedure consisting of an outer learning loop and an inner inference loop. During the outer loop, the parameters of the encoders, fusion network, decoders, and precision-prior network \(f(\mathbf{z})\) were optimized through gradient descent on variational free energy. The inner loop performed iterative inference over latent beliefs \(\mathbf{z}\) and modality precisions \(\mathbf{w}\) while network parameters remained fixed.
For each training sample, modality observations were first encoded through amortized inference using the modality-specific encoders and fusion network to obtain an initial latent state estimate \(\mathbf{z}_0\). The initial latent precision estimate \(\mathbf{w}_0\) was then obtained through the precision-prior network \(f(\mathbf{z}_0)\). These initial latent states were subsequently refined through iterative free-energy minimization in the inner loop.

The inner loop consisted of five inference iterations during training, while substantially larger numbers of iterations (typically 50--100) were used during test-time inference to analyze the convergence properties and arbitration dynamics of the model. 

To preserve stable encoder learning while preventing gradients from propagating through the iterative inference process, latent states refined during the inner loop were reattached to the original amortized latent states using a stop-gradient formulation:
\begin{equation}
    \begin{aligned}
        \mu_z& =\mu_{z,0}+(\mu_{z,\mathrm{refined}}-\mu_{z,0})_{\mathrm{detach}},\\
        \mu_w&=\mu_{w,0}+(\mu_{w,\mathrm{refined}}-\mu_{w,0})_{\mathrm{detach}}.
    \end{aligned}    
\end{equation}

This hybrid formulation combines efficient amortized initialization with iterative free-energy minimization, enabling latent states and modality precisions to adapt dynamically to sensory conflict, uncertainty, and modality dropout while avoiding unstable gradient propagation through multiple inference steps.

For training only, auxiliary regularization terms were added to encourage zero-centered precision latents and similar variance across precision dimensions, improving the interpretability of the learned precision-prior structure.
\begin{equation}
\mathcal{L}_{\mathrm{train}}
=
\mathcal{F}
+
\lambda_{\mathrm{center}}
L_{\mathrm{center}}
+
\lambda_{\mathrm{var}}
L_{\mathrm{var}},
\end{equation}

\subsection{Experimental Conditions}

To evaluate the contribution of dynamic precision inference and learned precision priors, we compared three model variants corresponding to progressively more expressive forms of multimodal inference:

\begin{enumerate}
    \item \textbf{Fixed-precision baseline:} Modality precisions were fixed and identical across all modalities with \(w_m = 0.0\).
    Inference was performed only over the latent state \(\mathbf{z}\), while the learned precision-prior network \(f(\mathbf{z})\) was disabled. This condition corresponds to standard multimodal latent inference without adaptive precision weighting.

    \item \textbf{Dynamic-precision model:} Modality precisions were inferred dynamically during iterative free-energy minimization, while the learned precision-prior network \(f(\mathbf{z})\) remained disabled. All modality precisions were initialized with \(w_m = 0.0\), but were free to adapt independently during inference. This condition isolates the contribution of dynamic precision inference alone.

    \item \textbf{Full model:} Both dynamic precision inference and the learned precision-prior network \(f(\mathbf{z})\) were enabled. In this setting, latent representations influenced the prior distribution over modality precisions, allowing learned class-dependent precision structures to shape multimodal arbitration dynamics.
\end{enumerate}

The proposed model variants were evaluated across two experimental settings designed to examine robustness under sensory corruption, multimodal inference from partial observations, and arbitration under inter-modal incongruence.

\subsubsection{Noise Robustness}

This experiment evaluated whether dynamic precision inference suppresses increasingly unreliable sensory modalities. For each test sample, one modality was corrupted using additive Gaussian noise:
\[
o_m^{\mathrm{noisy}}
=
o_m + \alpha \epsilon,
\qquad
\epsilon \sim \mathcal{N}(0, I),
\]
where \(\alpha \in [0.0,1.0]\) controlled the noise magnitude. Five noise levels were evaluated ranging from no corruption (\(\alpha=0.0\)) to strong corruption (\(\alpha=1.0\)).

For each experimental condition, 100 test samples were evaluated using 100 inner-loop inference iterations.

Modality precisions \(\mathbf{w}\) and modality-specific gradient contributions to latent-state updates of \(\mathbf{z}\) were analyzed to examine how precision inference modulates the influence of noisy sensory inputs.

Additionally, we examined how sensory corruption affects latent inference dynamics by comparing multimodal and unimodal inference across model variants. Latent robustness was quantified using latent trajectory deviation (LTD), defined as the average Euclidean distance between a perturbed inference trajectory and the corresponding clean inference trajectory across all inference steps. 

\subsubsection{Incongruency Arbitration}

This experiment evaluated whether modality-specific precisions actively guide latent belief formation under conflicting sensory evidence. Incongruent samples were constructed by combining modalities from two randomly selected samples \(A\) and \(B\), while the remaining modality was masked. For each trial, one modality provided observations from sample \(A\) and another from sample \(B\), producing conflicting multimodal evidence.

To avoid encoder bias toward higher-dimensional modalities, the latent state \(\mathbf{z}\) was initialized at the mean of the encoder-initialized latent states of samples \(A\) and \(B\). Initial modality precisions were then biased by a parameter \(\Delta w \in [-1,1]\), assigning opposite precision latents to the modalities originating from the two samples (\(+\Delta w\) for \(A\), \(-\Delta w\) for \(B\)).

Experiments were conducted on 100 randomly generated incongruent samples using 50 inner-loop inference iterations. Arbitration performance was evaluated using three measures: (1) arbitration accuracy, defined by the Euclidean distance of the final latent state to the latent encodings of samples \(A\) and \(B\), (2) latent trajectory length during inference, measuring convergence stability and directness in latent space, and (3) combined multimodal reconstruction error relative to both original samples, indicating overall reconstruction error across samples.

\section{Results}
\label{results}

\subsection{Learned Prior Structure}

The learned precision-prior network \(f(\mathbf{z})\) successfully acquired structured class-dependent precision representations. As shown in Fig.~\ref{fig:f(z)}, distinct latent categories occupy different regions of the precision space, reflecting modality-specific reliability expectations. Figure~\ref{fig:sep} shows pairwise distances between class centroids in \(\mathbf{w}\)-space. Classes sharing audio or tactile groupings remain separated despite their modality-level ambiguities, indicating that the learned prior encodes class-dependent reliability structure rather than modality-specific group assignments.

\subsection{Noise Robustness}

\subsubsection{Modality Suppression Under Noise}

This experiment examined the behavior of inferred modality precisions and modality-specific gradient contributions to latent-state updates under increasing sensory corruption. Results are shown for visual noise; the behavior was equivalent when corrupting the other modalities.

\begin{figure}[t]
    \centering
    \begin{subfigure}[b]{0.48\linewidth}
        \centering
        \includegraphics[width=\linewidth]{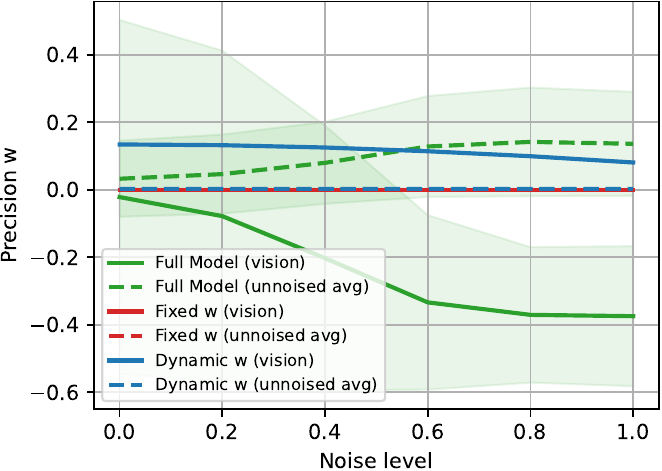}
        \caption{Analysis of latent \(\mathbf{w}\) dynamics.}
        \label{fig:precision}
    \end{subfigure}
    \hspace{0.3em}
    \begin{subfigure}[b]{0.48\linewidth}
        \centering
        \includegraphics[width=\linewidth]{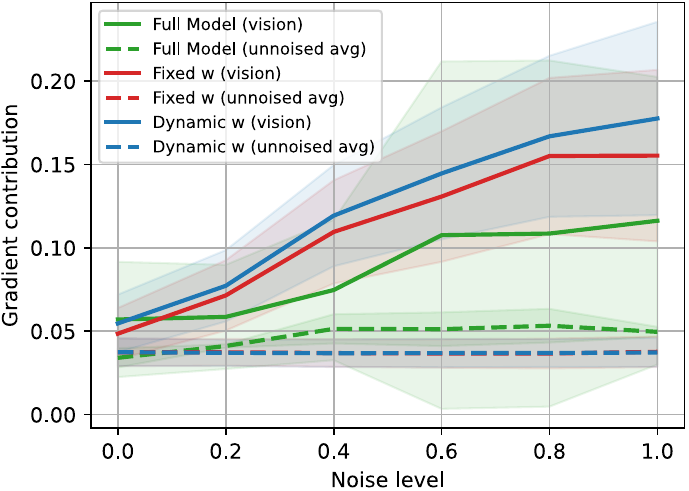}
        \caption{Analysis of gradient contributions.}
        \label{fig:grad}
    \end{subfigure}
    
    \caption{Noise robustness results for different levels of noise.}
    \label{fig:noise}
\end{figure}

As shown in Fig.~\ref{fig:precision}, the inferred precision of the corrupted modality decreases with increasing noise in both the dynamic-precision and full-model conditions. In the full model, the average precision of the uncorrupted modalities simultaneously increases, indicating adaptive cross-modal redistribution of inferential influence through the learned precision prior. In contrast, the fixed-precision baseline cannot adjust modality weighting during inference.

Figure~\ref{fig:grad} shows the modality-specific gradient contributions to latent-state updates. As visual noise increases, prediction-error gradients from the corrupted modality grow substantially in the baseline conditions. The full model suppresses these gradients through precision modulation, reducing the influence of noisy sensory evidence while increasing reliance on uncorrupted modalities.

\subsubsection{Latent Robustness}

To examine how sensory corruption influences latent belief dynamics, we compared inference trajectories obtained from noisy observations to those obtained from the corresponding clean observations. Robustness was quantified using latent trajectory deviation (LTD),

\begin{equation}
\mathrm{LTD}
=
\frac{1}{T}
\sum_{t=1}^{T}
\left\|
\mathbf{z}_t^{\mathrm{perturbed}}
-
\mathbf{z}_t^{\mathrm{clean}}
\right\|,
\end{equation}

which measures the average distance between perturbed and clean latent trajectories throughout inference. Lower LTD values indicate that latent belief dynamics remain closer to the clean inference process despite sensory corruption.

\begin{figure}[t]
\centering
\includegraphics[width=0.6\linewidth]{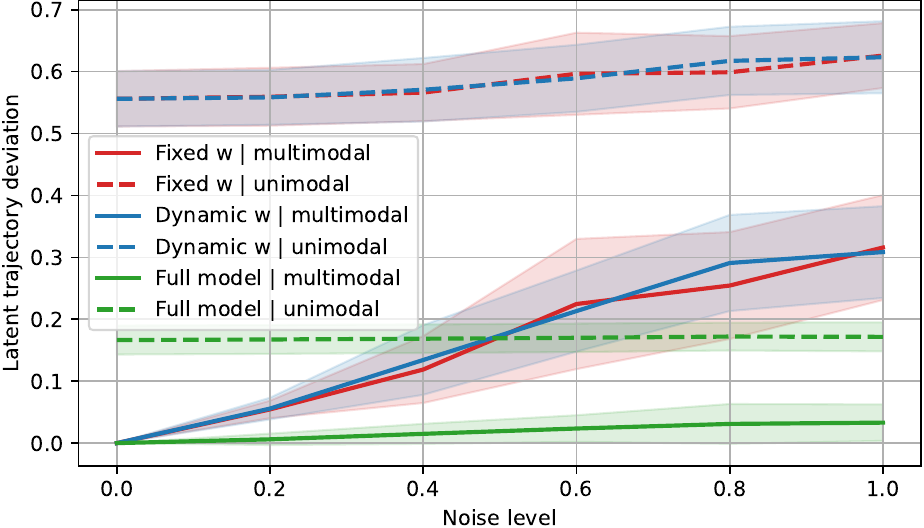}
\caption{Analysis of multimodal robustness to noise.}
\label{fig:robustness}
\end{figure}

Figure~\ref{fig:robustness} shows that unimodal inference produces substantially larger trajectory deviations than multimodal inference across all model variants, demonstrating the stabilizing effect of additional sensory cues under noisy conditions. Furthermore, the full model consistently exhibits the lowest LTD values, indicating that dynamic precision inference and learned precision priors improve robustness of latent belief dynamics to sensory corruption. This effect is consistent with the suppression of gradients from corrupted modalities and the increased inferential influence of uncorrupted sensory channels observed in the previous analysis.

\subsection{Incongruency Arbitration}

\begin{figure}[t]
    \centering

    \begin{subfigure}[b]{0.45\linewidth}
        \centering
        \includegraphics[width=\linewidth]{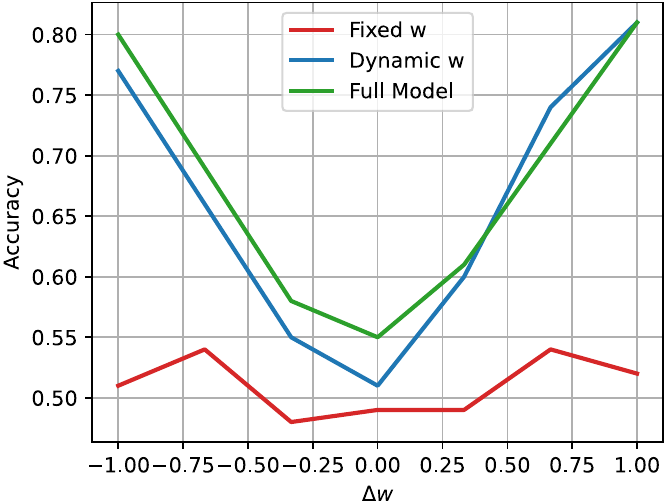}
        \caption{Arbitration accuracy.}
        \label{fig:a}
    \end{subfigure}
    \hspace{0.7em}
    \begin{subfigure}[b]{0.45\linewidth}
        \centering
        \includegraphics[width=\linewidth]{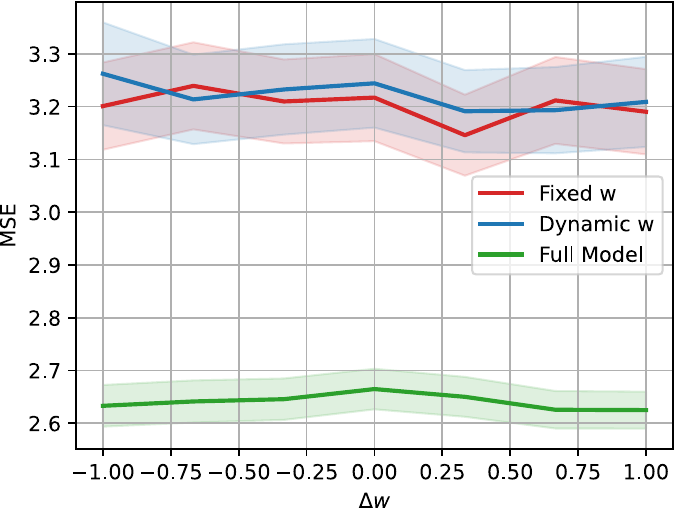}
        \caption{Reconstruction error sum.}
        \label{fig:b}
    \end{subfigure}

    \vspace{0.3em}

    \begin{subfigure}[b]{0.45\linewidth}
        \centering
        \includegraphics[width=\linewidth]{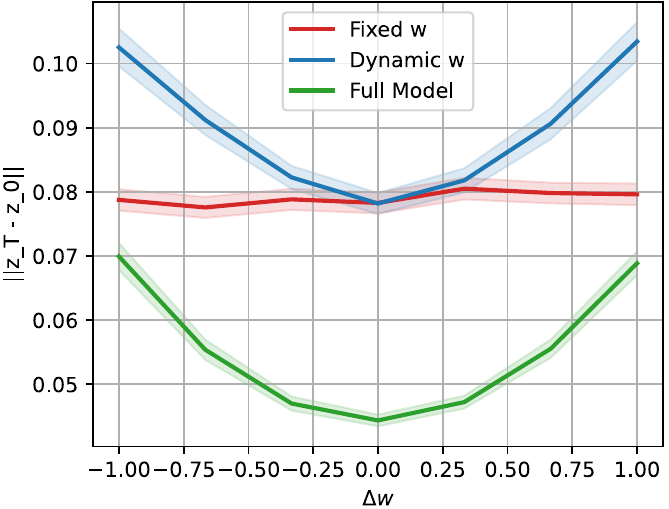}
        \caption{Latent trajectory length.}
        \label{fig:c}
    \end{subfigure}
    \hspace{0.7em}
    \begin{subfigure}[b]{0.45\linewidth}
        \centering
        \includegraphics[width=\linewidth]{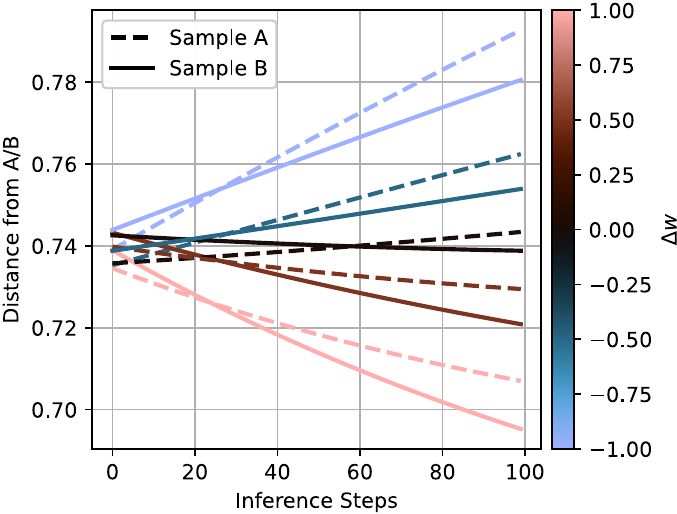}
        \caption{Single trial arbitration analysis.}
        \label{fig:d}
    \end{subfigure}

    \caption{Incongruency arbitration analysis.}
    \label{fig:main}
\end{figure}

Incongruent sensory inputs require arbitration between competing latent hypotheses. Figure~\ref{fig:a} shows that the fixed-precision baseline remains near chance level, whereas both dynamic-precision models achieve higher accuracy that increases with precision bias toward the target modality. The full model consistently outperforms the zero-prior variant and retains a modest advantage even at \(\Delta w=0\), indicating that learned precision priors contribute to arbitration beyond externally imposed precision biases.
This improvement is accompanied by lower reconstruction error (Fig.~\ref{fig:b}) and shorter latent trajectories (Fig.~\ref{fig:c}) compared to the baseline conditions, indicating more accurate and direct convergence toward a consistent latent state.

Figure~\ref{fig:d} illustrates a representative arbitration trial for the full model under different precision biases (\(\Delta w\)). Increasing precision for one modality causes the latent state z to converge toward the corresponding sample representation while moving away from the competing alternative. At \(\Delta w=0\), trajectory crossings are consistent with an influence of latent-dependent precision expectations encoded by \(f(\mathbf{z})\). These results show that precision shapes both sensory weighting and latent belief dynamics during multimodal conflict.

\section{Discussion}
\label{discussion}

The results demonstrate that precision inference successfully suppresses corrupted sensory channels and does not merely track uncertainty, but actively shapes latent belief dynamics. During multimodal conflict, changes in modality-specific precision enable successful arbitration between competing sensory hypotheses, supporting active inference accounts that interpret precision as a computational mechanism for attentional allocation. These findings provide evidence that precision-weighted prediction errors can support adaptive multimodal attention and conflict resolution.

Furthermore, learned precision priors induce class-dependent expectations regarding modality reliability. Different object categories acquire distinct modality importance structures that influence precision inference under noisy and incongruent sensory conditions. In this sense, the learned prior provides a latent-dependent precision mechanism in the form of top-down expectations about sensory reliability, analogous to attentional biases observed in biological systems.

Several limitations should be acknowledged. The experiments were conducted on a synthetic multimodal dataset, and the learned precision structures are therefore tied to artificially generated modality statistics. In addition, the proposed framework considers static observations rather than temporally evolving sensory streams, limiting its ability to model longer-term attentional dynamics. Future work will investigate more realistic multimodal environments, hierarchical precision representations, active sensory modality sampling, and precision formulations that more tightly couple latent beliefs and uncertainty estimation.

\section{Conclusion}
\label{conclusion}

This paper introduced a multimodal active inference framework that jointly infers latent beliefs and modality-specific sensory precisions during free-energy minimization. In addition, we proposed learned class-dependent precision priors that encode structured expectations regarding modality reliability. Through experiments involving sensory noise and multimodal incongruence, we demonstrated that dynamic precision inference improves robustness to sensory corruption, suppresses unreliable sensory evidence, and enables effective arbitration between competing sensory inputs.

The results suggest that sensory precision serves not only as an estimate of uncertainty, but also as a mechanism that shapes the evolution of latent beliefs during inference. Furthermore, learned precision priors provide top-down biases that influence multimodal arbitration and belief formation. Together, these findings highlight the potential of precision inference as a computational mechanism for adaptive and interpretable multimodal perception within active inference.

\begin{credits}
\subsubsection{\ackname} This research has been supported by JSPS grant JP23H04834.

\subsubsection{\discintname}
The authors have no competing interests to declare that
are relevant to the content of this article.
\end{credits}
%
%
%
\bibliographystyle{splncs04}
\bibliography{ref}
\end{document}